\PassOptionsToPackage{unicode}{hyperref}
\PassOptionsToPackage{hyphens}{url}
\PassOptionsToPackage{dvipsnames,svgnames,x11names}{xcolor}
\documentclass[
  11pt,
]{article}
\usepackage{xcolor}
\usepackage[margin=1in]{geometry}
\usepackage{amsmath,amssymb}
\usepackage{iftex}
\ifPDFTeX
  \usepackage[T1]{fontenc}
  \usepackage[utf8]{inputenc}
  \usepackage{textcomp} % provide euro and other symbols
\else % if luatex or xetex
  \usepackage{unicode-math} % this also loads fontspec
  \defaultfontfeatures{Scale=MatchLowercase}
  \defaultfontfeatures[\rmfamily]{Ligatures=TeX,Scale=1}
\fi
\usepackage{lmodern}
\ifPDFTeX\else
\fi
\IfFileExists{upquote.sty}{\usepackage{upquote}}{}
\IfFileExists{microtype.sty}{% use microtype if available
  \usepackage[]{microtype}
  \UseMicrotypeSet[protrusion]{basicmath} % disable protrusion for tt fonts
}{}
\makeatletter
\@ifundefined{KOMAClassName}{% if non-KOMA class
  \IfFileExists{parskip.sty}{%
    \usepackage{parskip}
  }{% else
    \setlength{\parindent}{0pt}
    \setlength{\parskip}{6pt plus 2pt minus 1pt}}
}{% if KOMA class
  \KOMAoptions{parskip=half}}
\makeatother
\usepackage{longtable,booktabs,array}
\usepackage{calc} % for calculating minipage widths
\usepackage{etoolbox}
\makeatletter
\patchcmd\longtable{\par}{\if@noskipsec\mbox{}\fi\par}{}{}
\makeatother
\IfFileExists{footnotehyper.sty}{\usepackage{footnotehyper}}{\usepackage{footnote}}
\makesavenoteenv{longtable}
\providecommand{\tightlist}{%
  \setlength{\itemsep}{0pt}\setlength{\parskip}{0pt}}
\DeclareUnicodeCharacter{2212}{\ensuremath{-}}
\DeclareUnicodeCharacter{00D7}{\ensuremath{\times}}
\usepackage{bookmark}
\IfFileExists{xurl.sty}{\usepackage{xurl}}{} % add URL line breaks if available
\makeatletter
\@ifundefined{xmpquote}{}{}
\makeatother
\hypersetup{
  pdftitle={Is the Judge Its Twin?},
  pdfauthor={Arman Nik Khah, The University of Texas at Dallas (arman.nikkhah@utdallas.edu); Arvin Bahreini, University of Oregon (arvinb@uoregon.edu)},
  colorlinks=true,
  linkcolor={Maroon},
  filecolor={Maroon},
  citecolor={Blue},
  urlcolor={Blue},
  pdfcreator={LaTeX via pandoc}}

\title{Is the Judge Its Twin?}
\usepackage{etoolbox}
\makeatletter
\providecommand{\subtitle}[1]{% add subtitle to \maketitle
  \apptocmd{\@title}{\par {\large #1 \par}}{}{}
}
\makeatother
\subtitle{Post-training makes a model's writing more predictable but
barely moves its taste, as a judge, toward predictable writing}
\author{Arman Nik Khah\\ The University of Texas at Dallas\\ \texttt{arman.nikkhah@utdallas.edu} \and Arvin Bahreini\\ University of Oregon\\ \texttt{arvinb@uoregon.edu}}
\date{September 2026}

\begin{document}
\maketitle

\subsection{Abstract}\label{abstract}

Language models now routinely grade other language models. If post-training makes a model's writing more predictable, it may also teach the same model, as a judge, to reward predictable writing. We test this directly. We follow two open 7-billion-parameter model families, OLMo-2 and Zephyr, from their base models through supervised fine-tuning (SFT) and direct preference optimization (DPO), and measure every stage as a writer of short stories and as a judge of story pairs. As writers, the models drift as feared. Each family's trained model finds its stages' stories more familiar in the order base, SFT, DPO, in all ten prompts, and a model from the other family finds the trained stories 7.0 to 9.2 percent (OLMo-2) and 24 to 25 percent (Zephyr) less surprising per token. As judges, they barely move toward predictable writing. Asked which story is better, a question every judge passes a competence check on, no trained judge's preference for the more predictable story grows by as much as one point of pick probability at its estimate. A one-sided 95\% upper bound on that growth is about 5 points on the widest pairs. Training instead makes judges more swayed by story length and by display order, and it breaks ``more creative'' as a question. Trained judges no longer reliably prefer a story to its own words in scrambled order. For these two families, the practical hazard in a small trained judge of creative writing lies in these measurable habits.

\subsection{1. Introduction}\label{introduction}

When a lab wants to know whether a new model writes better stories, it
increasingly asks another model. A model acting as a judge reads two
answers and says which is better. Its verdicts now rank models on public
leaderboards (Zheng et al., 2023), and the paradigm now has a survey
literature of its own (Gu et al., 2024). For tasks with a checkable
answer, a judge is a cheap stand-in for a human grader, proposed as a
way to scale evaluation (Hada et al., 2024). For creative writing there
is no checkable answer, so a sharper question appears. Whose taste is
the judge applying?

Post-training is the sequence of steps that turns a raw pretrained model
into an assistant. It is known to narrow a model's writing, so that
outputs converge on familiar shapes and familiar phrasing (Mohammadi,
2024; Zhang et al., 2025; Karouzos et al., 2026). The judge is usually a
post-trained model too, and it can come from the same family as the
writer. If the training that narrowed the writer also narrowed the
judge, the judge would reward the very predictability the writer drifted
toward. A model that learned to write more originally would then score
no better, and possibly worse. The judge would be the writer's twin.

The worry has partial support. Judges give higher scores than human
raters do to text with lower perplexity, that is, text the judge finds
easier to predict (Wataoka et al., 2024). A recent study finds that a
judge prefers the style typical of AI-written text over the
unpredictability of human-written stories (Tutone et al., 2026). Judges
also favor answers from models related to them (Li et al., 2025). What
no study has done is watch one model family's writing and judging change
together, stage by stage, through training. That is the direct test, and
public training checkpoints make it possible.

We ran it on two families that release a checkpoint after every training
step. OLMo-2 (Team OLMo, 2024) has four: the base model, supervised
fine-tuning (SFT, where the model imitates example answers), direct
preference optimization (DPO, where it learns from pairs of answers
ranked by preference; Rafailov et al., 2023), and a final stage that
adds reinforcement learning on tasks with checkable answers. Zephyr
(Tunstall et al., 2023) has three: the Mistral-7B base (Jiang et al.,
2023), SFT, and DPO. Every stage wrote short stories, and every stage
judged pairs of stories written by the other family.

\textbf{Finding 1: the writer drifts.} Every trained stage writes
stories that its own family's trained model finds more familiar than the
base model's stories, and base, SFT and DPO fall in that order in all
ten prompts in both families. A model from the other family also finds
the trained stories less surprising, by 7.0 to 9.2 percent per token for
OLMo-2 and 24 to 25 percent for Zephyr (Section 5).

\textbf{Finding 2: the judge barely moves toward predictable writing.}
Asked which story is better, the question every judge passes a
competence check on, no trained judge's preference for the more
predictable story grows by as much as one point of pick probability at
its estimate. OLMo-2's estimates edge the predicted way at DPO and
final; Zephyr's go either way depending on which model scores
predictability. Ten prompts leave room for more, so we bound it after
the fact: the one-sided 95\% upper bound on the growth is about 5 points
on the widest pairs, and 2.7 on an average pair. The
predicted drift is not detected, and the bound says how large it could
be. Asked which is more creative, no trained judge's estimate prefers
the more predictable story more than its base's does. The two families
do not agree on a direction, and we claim none (Section 6).

\textbf{Finding 3: training breaks the word ``creative''.} Asked which
story is more creative, every trained judge fails a basic competence
check that both base models pass. On scrambled text, some trained judges
stop discriminating and default to one answer slot. One, Zephyr's SFT
judge, prefers a story's words in scrambled order over the story itself
87\% of the time. Asked which story is better, every stage passes
(Section 7).

Together, the first two findings make the headline. Training changes how
these models write far more reliably than it changes how much they, as
judges, prefer predictable writing. The third finding points at a
different and more practical hazard. The risk in asking a trained judge
about creativity is not that it rewards the predictable. The risk is
that the question stops tracking whether the text is even a story.

Section 8 reports a follow-up that tried to remove the main design's
central confound, that each judged pair sets a story from one training
stage against a story from another, and could not. Its failure is
informative in its own right. In the Zephyr writer, the routes that made
stories less predictable also broke some of them, often enough that no
clean contrast survived a quality floor set in advance.

\subsection{2. Background}\label{background}

\textbf{Judges have known habits.} Model judges favor one answer
position regardless of content (Wang et al., 2023), prefer longer
answers, which length-controlled scoring was built to correct (Dubois et
al., 2024), and prefer their own writing (Panickssery et al., 2024).
Wataoka et al.~(2024) demonstrate that what looks like self-preference is a preference for
text the judge finds easy to predict, whoever wrote it. Bias taxonomies
now list this as a ``familiarity bias'' (Zhou et al., 2026). Tutone et
al.~(2026) show that a judge rating creative writing prefers AI-typical
style over human unpredictability. Li et al.~(2025) show that judges
favor generators related to them by training data or lineage. Each of
these studies examines finished models. None follows a judge through the
stages of its own training.

\textbf{Post-training narrows the writer.} Aligned models write text
that varies less, word by word and in overall shape, than their base
models' text (Mohammadi, 2024). Zhang et al.~(2025) trace this narrowing
partly to human preference data, which favors familiar text, and name it
typicality bias. Karouzos et al.~(2026) trace where diversity is lost
along several post-training paths and find the supervised fine-tuning
step to be a major site. This paper uses the same idea of typicality,
measured as predictability under a reader model (Section 3), on both
sides of the question.

\textbf{Predictability is not quality in either direction.} Human
writing does not sit at the most probable end of a language model's
distribution (Holtzman et al., 2020), so being hard to predict is not by
itself a defect. Yet text sampled to be improbable often is defective.
Raising the sampling temperature relates only weakly to judged novelty
and more strongly to incoherence (Peeperkorn et al., 2024), and the
quality of generated text falls at both ends of the likelihood range
(Zhang et al., 2020). Creativity is usually defined as novelty together
with value (Runco and Jaeger, 2012), so any test of a judge's taste for
the novel has to hold value fixed. Model judges of creative writing also
agree poorly with professional writers (Chakrabarty et al., 2024).

\subsection{3. Measuring how predictable a story
is}\label{measuring-how-predictable-a-story-is}

To ask whether a judge prefers predictable stories, we first need one
number for how predictable a story is. We use a reader model's surprise.
A reader is any language model; we feed it the story one token at a time
and record how much probability it gave to each token that actually came
next. The surprise at a token is the negative log of that probability,
measured in nats (natural-log units). A story's \textbf{surprise score}
is the average surprise per token. A lower score means a more
predictable story. This is the standard mean negative log-likelihood
(NLL), and it is our measure of what the literature calls typicality.

The score has a trap, and we fell into it before we fixed it. Scored
alone, the first few tokens of any story are expensive, because the
reader has no idea yet what the text is about. That cold-start cost is
spread over however many tokens follow, so short texts look
unpredictable just for being short. A seven-character OLMo-2 completion,
``Silence'', scored 6.2 nats per token this way. The fix is to put the
writing prompt in front of the story, exactly as the writer saw it, and
to average surprise over the story's tokens only. Every surprise score
in this paper uses that frame. We adopted it in a written amendment
before re-scoring, together with the two outcomes we expected (the
family-level drift of Section 5 would survive, and OLMo-2's drop under a
stranger reader would appear at about 7 to 10 percent); both
expectations were met.

We use the score in two ways to describe the writers, and in a third way
inside the judge analysis (Section 4).

\textbf{The stranger's view} asks whether a story is predictable to
anyone. The reader is the base model of the \emph{other} family, which
shares no training with the writer. OLMo-2 stories are read by the
Mistral-7B base; Zephyr stories are read by the OLMo-2 base.

\textbf{The family's view} asks whether training taught this family to
find this story familiar. For each story we take the score under the
family's fully trained model and subtract the score under the family's
own base model. A negative value means the trained model finds the story
easier to predict than its untrained ancestor did. Because both readers score the same text, a fixed offset between them
moves every stage by the same amount, so it cancels when stages are compared.

\subsection{4. Models, stories and
judges}\label{models-stories-and-judges}

\textbf{Writers and stories.} Seven checkpoints wrote stories:
OLMo-2-1124-7B at base, SFT, DPO and Instruct (the final stage), and
Mistral-7B-v0.1, mistral-7b-sft-beta and zephyr-7b-beta (Zephyr's DPO
and final stage). Each wrote 20 stories for each of ten flash-fiction
prompts (for example, ``Write a very short story about the last phone
call ever made''), in a bare completion frame
(\texttt{Writing\ prompt:\ \{prompt\}\textbackslash{}n\textbackslash{}nResponse:}),
at temperature 1.0 and top-p 0.95 (the usual settings for how freely a
model samples its next word), with at most 220 new tokens. We call each
family's last checkpoint its final stage, so Zephyr's DPO model is also
its final model. That is 1,400 generations. Eleven were empty, all from
OLMo-2, and were dropped, leaving 1,389 stories.

\textbf{Judges.} A judge must work at every stage, including base models
that cannot follow instructions. So we never ask a judge to write an
answer. We show it two stories and read the probability it assigns to
the next token, A or B, renormalized over those two letters, after this frame:

\begin{quote}
Here are two responses to the same writing prompt. Prompt: \{prompt\}

Response A: \{story\}

Response B: \{story\}

The more creative response is Response
\end{quote}

Each pair is shown in both orders and the two probabilities are
averaged, which removes a fixed slot preference from the measurement.
The preference itself remains a habit of the judge, and Section 7
measures it. The result, p\_x, is the judge's probability of choosing
story x. A second probe replaces ``creative'' with ``better''. In the
code that ran, the substitution produced the string ``The more better
response is Response''. Every ``better'' result in this paper used it, and
Section 7 re-checks it.

\textbf{A competence check before any judge is read.} A judge that
cannot do the task still produces probabilities, and a slope computed
from a non-judge means nothing. Worse, ``bias appears with training''
and ``competence appears with training'' would draw the same picture. So
before reading any judge's taste, we checked that it can tell a story
from a ruined copy of itself. For each prompt, six intact stories were
each paired with three ruins: the same words in random order
(\textbf{shuffled}), the first sentence alone (\textbf{stub}), and an
intact story written for a different prompt (\textbf{off-prompt}). That
gives 180 pairs per judge. We wrote the rule down before the competence
check ran. A judge passes if it prefers the intact story in at least
80\% of shuffled pairs and the lower end of the 95\% Wilson interval (a
standard confidence interval for a proportion) exceeds 50\%. Only the
shuffled test gates. Section 6 explains why the main measurement can
still read judges that fail this check under one wording.

\textbf{The main measurement.} Each judge stage then compared 200 pairs
of intact stories, 20 per prompt, all written by the other family, so in
this measurement no judge saw its own family's writing. The two stories
in a pair always come from different training stages of that family (a
base story against a DPO story, say), which matters in Section 6. In
plain terms, we ask how far a judge's pick moves toward a story as that
story gets more predictable, with length held fixed. For each judge and
criterion we fit

p\_x = a + b × (difference in surprise score, x minus y) + c ×
(difference in log length) ,

where each story's surprise score comes from its own family's base model
(the judge, recall, is always from the other family). A negative b means
the judge prefers the more predictable story. We fit it three ways. The
first is the model as written. The second adds the judge's own
``better'' probability as a covariate, which asks whether any
predictability preference survives once the judge's own quality call is
held fixed. The third uses the creativity-specific contrast, p\_creative
minus p\_better, as the outcome. The unit of inference is the prompt, so
every interval is a 95\% cluster bootstrap over the ten prompts, which
resamples whole prompts because stories written for one prompt are not
independent.

\textbf{Rules fixed before their runs.} Two rules
were recorded in a dated design document, with their expected outcomes,
before the runs that tested them: the prompt-in-front scoring and the
competence rule with its 80\% bar. The competence word was switched from
``creative'' to ``better'' after a pilot showed ``better'' working where
``creative'' failed, and before every run the check gates. The same
amendment set the judging limits used in Section 6.

\subsection{5. The writer drifts}\label{the-writer-drifts}

\textbf{Table 1.} How predictable each stage's stories are. Stranger's
view column: average surprise per token under the other family's base
model, with the base row in nats and later rows as the percent change
from base. Family's view column, in nats per token for every row: score
under the family's trained model minus score under its base model;
negative means the trained model finds the story more familiar.

{\def\LTcaptype{none} % do not increment counter
\begin{longtable}[]{@{}
  >{\raggedright\arraybackslash}p{(\linewidth - 6\tabcolsep) * \real{0.2500}}
  >{\raggedright\arraybackslash}p{(\linewidth - 6\tabcolsep) * \real{0.2500}}
  >{\raggedright\arraybackslash}p{(\linewidth - 6\tabcolsep) * \real{0.2500}}
  >{\raggedright\arraybackslash}p{(\linewidth - 6\tabcolsep) * \real{0.2500}}@{}}
\toprule\noalign{}
\begin{minipage}[b]{\linewidth}\raggedright
Family
\end{minipage} & \begin{minipage}[b]{\linewidth}\raggedright
Stage that wrote the story
\end{minipage} & \begin{minipage}[b]{\linewidth}\raggedright
Stranger's view
\end{minipage} & \begin{minipage}[b]{\linewidth}\raggedright
Family's view
\end{minipage} \\
\midrule\noalign{}
\endhead
\bottomrule\noalign{}
\endlastfoot
OLMo-2 & base & 1.915 & +0.434 \\
& SFT & −9.2\% & −0.121 \\
& DPO & −7.0\% & −0.567 \\
& final & −7.7\% & −0.581 \\
Zephyr & base & 2.105 & +0.243 \\
& SFT & −24.9\% & −0.139 \\
& DPO (final) & −24.0\% & −0.325 \\
\end{longtable}
}

Begin with the family's view, because it asks the question the twin
worry turns on. Did training teach this family to find its own kind of
writing familiar? It did. OLMo-2's trained model finds its base model's
stories harder to predict than the base model itself does (+0.434), and
finds each later stage's stories progressively easier (−0.121, −0.567,
−0.581). Zephyr follows the same order (+0.243, −0.139, −0.325). The
base stage is the most foreign in all ten prompts, and the order base,
then SFT, then DPO holds inside every one of the ten prompts in both
families. OLMo-2's final stage is indistinguishable from its DPO stage
(−0.581 against −0.567). From base to DPO the change is −1.000 nats per
token for OLMo-2 (95\% interval −1.06 to −0.95) and −0.569 for Zephyr
(−0.59 to −0.55).

One caution applies at the two ends of the order. A base-stage story is
scored partly by the model that wrote it, and so is a final-stage story,
and a model finds its own samples easy to predict. That pushes the two
end stages apart by construction. The middle of the order is the cleaner
evidence. OLMo-2's SFT and DPO stories were written by neither of their
scorers and still fall in order (−0.121, then −0.567), and the
stranger's view below involves no self-scoring at all. Zephyr has only
one middle stage, because its DPO model is also its final model, so its
order leans more on the ends. The family-specific breakdown in the next
paragraph inherits the same caution at its ends; its middle stages also
fall in order (OLMo-2 SFT −0.264, DPO −0.542).

A second objection is that this is just generic chatbot style, in which
case any heavily trained model would find any trained model's prose
familiar. Part of it is. The other family's trained model, asked the
same question about the same stories, agrees story by story (correlation
0.68 on OLMo-2 stories, 0.80 on Zephyr's). So we subtracted the other family's
view of each story from its own family's view. What remains is family-specific, and it still
orders the stages. It spans 0.692 nats per token for OLMo-2 and 0.261
for Zephyr, with base the most foreign in all ten prompts and the
base-to-DPO drop present in all ten (sign test p = .002). Much of the
familiarity that varies from story to story is shared chatbot style. How
much of the change across stages is family-specific differs by family.
Measured as the spread between the least and the most familiar stage,
for OLMo-2 it is most of the change (0.692 of 1.015, the spread from
base to final), and for Zephyr a little under half (0.261 of 0.569).

The stranger's view asks a blunter question, whether the writing becomes
predictable to anyone. It does, and here the drop arrives almost
entirely at SFT; later stages hold it rather than extend it. For Zephyr
the drop is large. The OLMo-2 base model finds Zephyr's SFT stories
24.9\% less surprising per token than Zephyr's base stories (2.105 falls
to 1.581) and its DPO stories 24.0\% less, with the drop in all ten
prompts (sign test p = .002 at each stage). OLMo-2's drift is smaller:
9.2\%, 7.0\% and 7.7\% at SFT, DPO and final, read by the Mistral base.
It appears in 9, 9 and 8 of the ten prompts, and every cluster-bootstrap
interval excludes zero (in nats per token, −0.178 {[}−0.248, −0.102{]},
−0.134 {[}−0.191, −0.064{]} and −0.148 {[}−0.205, −0.084{]}).

The two views part company at DPO. The stranger gives back a sliver
(OLMo-2 from −9.2\% to −7.0\%, Zephyr from −24.9\% to −24.0\%), while
the family's own trained model keeps finding the writing more familiar
(OLMo-2 from −0.121 to −0.567, Zephyr from −0.139 to −0.325).

The generator side of the twin hypothesis therefore holds in both
families. Training makes the writing more predictable, and it makes each
family's trained model find its own family's writing familiar.

\subsection{6. The judge barely moves toward predictable
writing}\label{the-judge-barely-moves-toward-predictable-writing}

\textbf{Table 2.} How much each judge's preference depends on
predictability. Slope b from Section 4, per judge stage, 200 pairs each.
A negative slope means the judge prefers the more predictable story.
Brackets: 95\% cluster-bootstrap interval over prompts; the ``holding
`better' fixed'' column is a point estimate only. The first column is
the reading this section leads with; the other three are read under
``creative''.

{\def\LTcaptype{none} % do not increment counter
\begin{longtable}[]{@{}
  >{\raggedright\arraybackslash}p{(\linewidth - 8\tabcolsep) * \real{0.170}}
  >{\raggedright\arraybackslash}p{(\linewidth - 8\tabcolsep) * \real{0.220}}
  >{\raggedright\arraybackslash}p{(\linewidth - 8\tabcolsep) * \real{0.220}}
  >{\raggedright\arraybackslash}p{(\linewidth - 8\tabcolsep) * \real{0.170}}
  >{\raggedright\arraybackslash}p{(\linewidth - 8\tabcolsep) * \real{0.220}}@{}}
\toprule\noalign{}
\begin{minipage}[b]{\linewidth}\raggedright
Judge
\end{minipage} & \begin{minipage}[b]{\linewidth}\raggedright
``better'' slope \mbox{{[}95\% CI{]}}
\end{minipage} & \begin{minipage}[b]{\linewidth}\raggedright
``creative'' slope \mbox{{[}95\% CI{]}}
\end{minipage} & \begin{minipage}[b]{\linewidth}\raggedright
``creative'', holding the judge's own ``better'' fixed
\end{minipage} & \begin{minipage}[b]{\linewidth}\raggedright
``creative'' minus ``better''
\end{minipage} \\
\midrule\noalign{}
\endhead
\bottomrule\noalign{}
\endlastfoot
OLMo-2 base & \mbox{−.057 {[}−.12, −.01{]}} & \mbox{−.096 {[}−.18, −.02{]}} & −.022 &
\mbox{−.039 {[}−.07, −.01{]}} \\
OLMo-2 SFT & \mbox{−.048 {[}−.11, +.01{]}} & \mbox{−.058 {[}−.16, +.03{]}} & +.001 &
\mbox{−.010 {[}−.05, +.02{]}} \\
OLMo-2 DPO & \mbox{−.064 {[}−.16, +.03{]}} & \mbox{−.051 {[}−.19, +.07{]}} & +.019 &
\mbox{+.013 {[}−.04, +.06{]}} \\
OLMo-2 final & \mbox{−.069 {[}−.18, +.03{]}} & \mbox{−.054 {[}−.21, +.08{]}} & +.020 &
\mbox{+.016 {[}−.03, +.06{]}} \\
Zephyr base & \mbox{+.024 {[}−.01, +.06{]}} & \mbox{+.040 {[}−.01, +.10{]}} & +.007 &
\mbox{+.016 {[}−.01, +.04{]}} \\
Zephyr SFT & \mbox{+.023 {[}−.02, +.08{]}} & \mbox{+.052 {[}−.00, +.11{]}} & +.021 &
\mbox{+.029 {[}+.01, +.05{]}} \\
Zephyr DPO & \mbox{+.076 {[}−.03, +.20{]}} & \mbox{+.120 {[}+.03, +.24{]}} & +.046 &
\mbox{+.044 {[}−.01, +.11{]}} \\
\end{longtable}
}

\textbf{The reading we lead with: ``better''.} Every judge passes the
competence check when asked which story is better (Section 7), so these
are the slopes a reader can trust most. If the judge were the writer's
twin, they would be negative and would grow more negative from base to
final. OLMo-2's slopes are negative at every stage. They edge the
predicted way at DPO and final (−.064 and −.069, against −.057 at base),
after a smaller lean at SFT (−.048). Zephyr's slopes sit near zero at
base and SFT (+.024, +.023) and point away from the predictable story at
DPO (+.076). Only OLMo-2's base interval excludes zero.

\textbf{How much drift the data allow.} Ten prompts give wide intervals,
so we asked directly how far each trained judge's slope could have moved
toward the predictable story beyond its own base judge's.\footnote{The two analyses labelled post hoc, the drift bound here and the stage-difference control below, share the main analysis's resampling design.} Every stage of a
family judged the same 200 pairs, so we resampled prompts once per draw
and refit the base judge and the trained judge on the same draw (5,000
draws), and took the one-sided 95\% upper bound on drift toward the
predictable story. We ran it twice, once with the story-family reader
(each story's surprise scored by its own family's base model, as in
Table 2) and once with the judge-family reader (the judge's family base
model, which involves no self-scoring). Under ``better'', the bounds are
.011, .051 and .064 per nat for OLMo-2's SFT, DPO and final judges with
the story-family reader (.013, .048 and .060 with the judge-family
reader), and .019 and .016 for Zephyr's SFT and DPO judges (.029 and
.069 with the judge-family reader). To put that in pick terms, the
average judged pair differs in surprise score by 0.40 nats per token for
the stories OLMo-2's judges read (0.45 with the judge-family reader) and
0.33 for Zephyr's (0.30). In pick terms the largest bound is OLMo-2's
final judge's: its pick probability could move toward the predictable
story by up to 2.7 points with the judge-family reader (2.6 with the
story-family reader). For Zephyr the largest is its DPO judge's, 2.1
points with the judge-family reader (0.5 with the story-family reader).
For scale, OLMo-2's base judge's whole lean is about 2.6 points with the
judge-family reader (−.059 × 0.45) and 2.3 with the story-family reader
(−.057 × 0.40), so the data cannot exclude that lean roughly doubling.
At the p90 pair of story difference, .82 nats per token for OLMo-2 and
.62 for Zephyr, the bounds are 4.9 and 4.3 points, and at the p95 pair,
.91 and .75 nats, 5.5 and 5.2, all with the judge-family reader. The
bound covers movement toward the predictable story only.

\textbf{The second reading: ``creative''.} Trained judges fail the
competence check under ``creative'', so are these slopes worth reading
at all? On the real story pairs they carry information. The
scrambled-text failure shows up on word salad, and real pairs contain
none. On these 200 pairs every judge's picks vary (standard deviation of
p\_x from .14 to .34 under ``creative'') and disagree across display
orders by .12 to .38 on average, inside both limits set in advance (.07
and .45). Two
real pairs show what that means for Zephyr's SFT judge, the one that
preferred word salad. For the museum prompt, it compared a base-model
story that repeats its own prompt and stops mid-sentence (``I wish'')
with a final-stage story about a silver pitcher of similar length (878
and 817 characters). It picked the pitcher story as more creative with
probability .87, in both display orders. For the red-light prompt, it
preferred a story whose driver runs the light, and which ends in a
leaked ``Word count: 153'', over one that stays at the light (.88; 564
and 649 characters). We chose these two to illustrate, one sensible pick
and one not. On real text these judges produce varied picks within the
limits set in advance; whether they pick well is a separate question,
and that is why ``creative'' is the second reading, not the first.

OLMo-2's judges' estimates lean toward the more predictable story when
asked which is more creative. The lean is largest in the untrained base
model (−.096, the only OLMo-2 interval that excludes zero), and it is
smaller after training (−.058, −.051, −.054). Training does not add to
it. Zephyr's estimates point the other way, toward the less predictable
story, and grow with training, from +.040 at base to +.052 after SFT and
+.120 after DPO; only the DPO interval excludes zero. The Zephyr pattern
survives dropping every pair that contains a story under 400 characters
(+.051, +.071, +.112). Under ``creative'' the drift bounds are smaller
still: with the story-family reader, .016 and .026 for OLMo-2's DPO and
final judges and below zero for the other three, and with the
judge-family reader at most .035 for any trained judge.

Holding the judge's own ``better'' verdict fixed shrinks every
``creative'' slope to within .05 of zero. The creativity-specific
contrast is small everywhere, from −.039 to +.044, and it does not move
toward predictability with training in either family. If anything it moves away, rising at every stage in both families (Table 2), a direction this design cannot separate from stage. Two of its
intervals exclude zero, and they point in opposite directions. OLMo-2's
base judge leans toward the predictable story and Zephyr's SFT judge
away from it.

We also re-ran every fit two ways: on long stories only (at least 400
characters, a sensitivity cut set in advance), and with surprise scored
by the judge's own family base model instead of the story's, a second
sensitivity check (for the stories a judge reads, that is the
stranger's-view reader of Section 3). Under ``creative'', no form shows a trained judge
preferring the predictable story more than its base does. Under
``better'', some trained judges drift the predicted way in some forms.
The largest such drift, .016, is Zephyr's DPO judge with surprise scored
by the judge's own family (−.016 against −.000 at base), well inside its
interval. The predicted pattern, a clear preference for predictable
writing that appears or strengthens with training, is not detected.

\textbf{The stage control.} In this design every pair sets one training
stage's story against another's. Because training makes stories more
predictable (Section 5), the predictability difference within a pair is
nearly the same thing as the stage difference. We added the stage difference as a covariate. It was run only on the long-story
subset (both stories at least 400 characters), not on all pairs; for
OLMo-2's judges that subset is every pair, and for Zephyr's it is 152 of
200. There, under ``creative'', it moves OLMo-2's slopes from −.028 at
base to +.026, +.072 and +.079 for SFT, DPO and final, all intervals
including zero. Under ``better'' it moves them from −.032 at base to
−.011, +.009 and +.011. Zephyr's ``creative'' slopes stay positive under
the same control (+.044, +.061, +.091). When one nearly redundant
covariate flips a sign, the data cannot separate ``rewards less
predictable writing'' from ``rewards later-stage writing''. This paper
therefore claims neither sign for the judge slope. It claims only that
the predicted drift is not detected, and bounds how large it could be.
Separating the two readings needs pairs from one writer at one stage
that differ in predictability alone, which is what Section 8 attempted.

\subsection{7. What training does to a judge
instead}\label{what-training-does-to-a-judge-instead}

The judges are not inert. On the same pairs, the spread of their picks
grows from base to final (standard deviation of p\_x under ``creative''
from .14 to .25 for OLMo-2 and from .16 to .34 for Zephyr), and several
of their habits grow.

\textbf{The word ``creative'' stops working.} The competence check told
us this first (Table 3). Asked for the more creative story, both base
models pass, preferring the intact story over its shuffled copy 95\%
(OLMo-2) and 88\% (Zephyr) of the time. Every trained stage fails, in
one of two ways. OLMo-2's three trained stages and Zephyr's DPO stage
mostly answer ``A'' in both display orders, so they sit near chance.
They default to a slot rather than judge. Zephyr's SFT stage does
something stranger. It picks the scrambled text 87\% of the time, with
99\% of its next-token probability on the two answer letters. It is
answering, confidently, and its answer is word salad.

\textbf{Table 3.} Competence check, share of shuffled pairs in which the
judge prefers the intact story (60 pairs per judge). Pass requires at
least .80 and a Wilson lower bound above .50.

{\def\LTcaptype{none} % do not increment counter
\begin{longtable}[]{@{}lll@{}}
\toprule\noalign{}
Judge & asked ``more creative'' & asked ``better'' \\
\midrule\noalign{}
\endhead
\bottomrule\noalign{}
\endlastfoot
OLMo-2 base & .95 & 1.00 \\
OLMo-2 SFT & .47 & .97 \\
OLMo-2 DPO & .47 & .98 \\
OLMo-2 final & .48 & .98 \\
Zephyr base & .88 & .97 \\
Zephyr SFT & .13 & .88 \\
Zephyr DPO & .40 & .98 \\
\end{longtable}
}

Asked for the better story, every stage passes. To check that the word,
not the formatting, was responsible, we crossed the criterion with the
prompt format on the two SFT judges. Wrapping the question in each
model's chat template changed nothing. Changing the word changed
everything. OLMo-2 SFT went from .47 (bare frame) and .40 (chat
template) under ``creative'' to .97 and 1.00 under ``better'', and
Zephyr SFT went from .13 and .07 to .88 and .88. Because the ``better''
probe as run was ungrammatical, we repeated the competence check, after
seeing the results, on both SFT judges with the grammatical wording.
Both pass. OLMo-2 SFT prefers the intact story in .95 of shuffled pairs
and Zephyr SFT in .97, against .97 and .88 with the original wording.
The typo is not what rescues the judges. Our reading of Zephyr's SFT
judge is that it takes ``creative'' to mean ``unusual'', with no floor
on value, and scrambled text is maximally unusual. The other four
trained judges mostly fall back on a slot, though Zephyr's DPO judge
still picks the scrambled text 60\% of the time.

This failure runs in the opposite direction from the twin hypothesis. A
twin judge would punish the unusual. Zephyr's SFT judge, asked about
creativity, rewards the unusual without regard to whether the text is a
story at all, and on scrambled pairs the other trained judges stop
discriminating. We report it as a finding of its own (Finding 3) because
it is the more immediate hazard for anyone using a small trained model
to rate creative writing. The question ``which is more creative?'' can
produce confident answers that no longer track anything a reader would
call creative.

Two other checks were reported for every stage. Stub pairs were easy,
and every judge preferred the full story over its first sentence at
least 95\% of the time. Off-prompt pairs were not. Accuracy stayed
between .50 and .70 under ``better'', so no judge reliably notices that
a story answers a different prompt. Our claims are therefore about the
text of a story, not about how well it fits its prompt.

\textbf{Length and position preferences grow.} Under ``creative'', the
slope on length difference rises with training, from +.005 to +.078,
+.102 and +.108 for OLMo-2, and from +.121 to +.131 and +.202 for
Zephyr, so longer stories win more often. Under ``better'' the OLMo-2
length slope moves the other way, from −.069 at base to −.158 at the
final stage, while Zephyr's still favors longer stories and grows
(+.071, +.090, +.175). The judges' preference for one display slot also
rises under ``creative'' from base to final, from .15 to .37 for OLMo-2
and from .15 to .21 for Zephyr (mean disagreement between the two
display orders), though not at every step (Zephyr's SFT judge dips to
.12, and OLMo-2's peaks at DPO, .38). These are the habits a
practitioner should correct for. These data show no typicality preference
added by training that would need its own correction.

\subsection{8. Manipulating predictability degrades quality}\label{why-a-cleaner-test-could-not-be-built}

Section 6 left one door open. Because every pair in the main design
mixes two training stages, a preference for less predictable writing and
a preference for later-stage writing cannot be told apart. The clean
experiment is easy to state. Take one writer at one stage. Produce pairs
of stories for the same prompt, of similar length and similar quality,
that differ in predictability. Ask judges from every stage which they
prefer. If later judges increasingly prefer the more predictable story,
the twin reading comes back.

We ran the first half of that experiment, building the pairs, with
Zephyr's final model (zephyr-7b-beta) as the writer and twelve
flash-fiction prompts. This time stories were allowed to finish. The
token limit was raised to 768, and all 49 ordinary stories (four per
prompt, five for one) ended on their own, the longest at 681 tokens; the
asking route was rerun at 1,024 so that its longer stories could finish
too. Rules set at the start of this work, and applied mechanically
before any story was read or scored, kept only finished stories (a
natural stop, final punctuation, at least 100 words, no leaked template
text) and paired stories only when the longer was at most 1.25 times the
length of the shorter.

Two gates stood between the pairs and any judge. The first is a
\textbf{manipulation check}, which requires the treated side of each
pair to be measurably less predictable to an outside reader (OLMo-2's
base model, prompt in front). The second is a \textbf{blind read}, in
which the two authors scored every story in four
sample pairs independently and resolved disagreements by discussion into one score per
story, from 1 to 5 for being a coherent,
complete story that fits its prompt, and recorded jointly which story in
each pair was better and which was more creative. The key to which story
was which was kept in a separate file that neither rater saw, and the
agreed ratings were committed to version control before it was opened.
The pass rule was written before the ratings were recorded or unblinded.
Every treated story must score at least 3, and the treated stories'
median must be at least 4.

We tried three ways of producing the treated side (Table 4). Predictability and judged quality moved together. The raters preferred the predictable story in 6 of 8 pairs, and both routes that lowered predictability degraded it enough to fail the blind read's floor. Asking did not move predictability. Heating and selecting did, and each produced at least one broken story.

\textbf{Table 4.} Three routes to a less predictable story from the same
writer. Treated side: the story the route was meant to make less
predictable (for selecting, the less predictable story of a pair chosen
by natural spread). In one heated pair it came out more predictable (see
below the table). Gap: average surprise per token, treated side minus
partner, under the OLMo-2 base reader, with a 95\% bootstrap interval.
Blind read: coherence scores (1 to 5) for the four sampled pairs.

{\def\LTcaptype{none} % do not increment counter
\begin{longtable}[]{@{}
  >{\raggedright\arraybackslash}p{(\linewidth - 8\tabcolsep) * \real{0.110}}
  >{\raggedright\arraybackslash}p{(\linewidth - 8\tabcolsep) * \real{0.250}}
  >{\raggedright\arraybackslash}p{(\linewidth - 8\tabcolsep) * \real{0.300}}
  >{\raggedright\arraybackslash}p{(\linewidth - 8\tabcolsep) * \real{0.170}}
  >{\raggedright\arraybackslash}p{(\linewidth - 8\tabcolsep) * \real{0.170}}@{}}
\toprule\noalign{}
\begin{minipage}[b]{\linewidth}\raggedright
Route
\end{minipage} & \begin{minipage}[b]{\linewidth}\raggedright
How the treated story was made
\end{minipage} & \begin{minipage}[b]{\linewidth}\raggedright
Gap in surprise (nats per token)
\end{minipage} & \begin{minipage}[b]{\linewidth}\raggedright
Blind read, treated side
\end{minipage} & \begin{minipage}[b]{\linewidth}\raggedright
Blind read, partner
\end{minipage} \\
\midrule\noalign{}
\endhead
\bottomrule\noalign{}
\endlastfoot
Asking & an added instruction to write a response with under 10 percent
probability & \mbox{−0.031 {[}−0.097, +0.037{]}}; 11 of 29 pairs in the
intended direction & not run & not run \\
Heating & sampling temperature raised from 1.0 to 1.3 & \mbox{+0.768 {[}+0.595, +0.908{]}}; 19 of 20 pairs & 2, 3, 4, 4 & 5, 5, 5, 5 \\
Selecting & ordinary stories at temperature 1.0, paired by natural
spread & \mbox{+0.176 {[}+0.053, +0.302{]}}, measured on held-out tokens & 4.5,
1, 2, 4 & 2, 3.5, 4, 3 \\
\end{longtable}
}

In one of the four heated pairs read blind, the heated story (scored 3)
was in fact the more predictable of the two (1.355 against 1.710 nats
per token).

\textbf{Asking did not move predictability.} Told to write a story it
would give under 10 percent probability, the writer produced stories
that were, if anything, slightly more predictable, and longer. The
reader can see real differences. With the same frame it had separated
Zephyr's training stages in the main study (Section 5), and it registers
the heating route's shift below. Our reading is that, asked for the
improbable, the writer reached for melodrama (apocalypses, last calls,
hidden rooms), which may be exactly what a base model expects.

\textbf{Heating moved predictability and broke a story.} At temperature
1.3 the gap was large and consistent. The blind read failed both legs of
its rule, because one heated story scored 2 and the median was 3.5. That
story contradicts itself. Its map shows a river, the next line says the river
``hasn't existed for centuries'' and asks how it can be ``missing from your map'',
and a paragraph later it ``must've gotten erased''. The raters
preferred the ordinary story as better and as more creative in all four
pairs.

\textbf{Selecting gave a real but smaller spread, and its widest pairs
failed the same floor.} Ordinary stories for one prompt already vary in
predictability. Measuring that spread fairly takes care, because if
pairs are chosen for having the widest gap, the choosing inflates the
gap. So we chose pairs using one half of each story's tokens and
measured the gap on the other half. The held-out gap was +0.176 nats per
token, with 7 of 11 prompts positive (one prompt had no pair within the
length rule). About a third of the naive gap was selection, since the
same pairs measured on the tokens used to choose them gave 0.260. The
interval overstates our confidence. On synthetic banks of pure noise,
the same procedure declared a gap real 9 times in 150 (6\%, where the
lower end of a two-sided 95\% interval should clear zero by chance about
2.5\% of the time). Roughly, since the quantities are measured
differently, the natural spread is a quarter of the temperature gap and
a third of the drift from Zephyr's base to its SFT stage. For the blind
read we deliberately took the widest pair from each of four prompts as a
stress test, since if quality falls with predictability, it should show
most there. Two of the four less predictable stories scored 1 and 2.
Both contain logic breaks. In one, an inherited debt turns into a
thriving business, and in the other a clock is a neon readout in one
sentence and a metal ball in the next. The other two scored 4 and 4.5,
and the raters picked the 4.5 as both better and more creative. The one
low score on the predictable side (a 2) reads fluently and plausibly
misses its prompt. None of the eight stories repeats itself (the share
of three-word sequences that repeat an earlier one is at most 0.032 in
each), so this is not the high-likelihood repetition failure that Zhang
et al.~(2020) describe.

\textbf{Both reads together.} Counting across both reads by each pair's
measured predictability rather than by route, three of the eight less
predictable stories scored 2 or below, all with logic breaks, against
one of the eight more predictable stories. Median coherence was about
the same on the two sides (4 on the less predictable side, 3.75 on the
more predictable). The raters picked the more predictable story as
better in 6 of 8 pairs and as more creative in 6 of 8. These are agreed
scores from one pair of raters on eight pairs, four of them chosen to be
extreme, so we report counts, not a test. What the counts do show is
that neither route delivered a set of less predictable stories that
reliably held together, which is what the floor demanded.

Appendix A checks whether the reader's surprise falls on the broken
sentences themselves. In the three broken stories it does not
concentrate there.

\textbf{What this means for the judge question.} The question Section 6
left open, whether a judge prefers the more predictable of two stories
when quality is held fixed, could not be asked with the pairs we could
build, because in them predictability and quality did not come apart
cleanly. Some less predictable stories broke, and the two human raters
mostly preferred the more predictable story. A judge that did the same
on these pairs might be making the same defensible call. No judge was
run on any of these pairs. Answering the question needs less predictable
stories that reliably hold together, from a stronger writer or from
human-written stories with human quality ratings. Temperature is a poor
lever for this, in line with Peeperkorn et al.~(2024).

\subsection{9. Discussion}\label{discussion}

The twin worry is reasonable, and on the writer's side it is borne out.
Post-training narrows these writers, and each family comes to find its
own narrowed writing familiar. The same training barely moves these
models, as judges, toward predictable writing, and a one-sided bound
says how far it could have. What training does teach them is a set of
habits already known from larger judges: favoring longer stories (under
``creative'' in both families, and under ``better'' in Zephyr), favoring
a slot, and, most sharply, losing any stable meaning for ``more
creative''. So for these two families, the practical risk in using a
small trained model to judge creative writing is not a hidden taste for
predictability. It is the ordinary, measurable habits, plus a criterion
word that no longer means what the person asking intends.

The follow-up adds a second lesson, about measurement. A surprise score
is often read as a proxy for novelty. In this writer, both routes that
lowered predictability also produced broken stories, often enough to
fail a quality floor set in advance. A study that manipulates
predictability and then asks whether judges reward it has to show,
independently, that its less predictable items are still good. Otherwise
a judge that ``penalizes the unpredictable'' may simply be penalizing
the incoherent, and it would be right to.

\subsubsection{Limitations}\label{limitations}

\begin{itemize}
\tightlist
\item
  \textbf{Scale.} Two families at 7B parameters, one size each. The
  planned third family (Tülu-3, built on Llama-3.1) was not run.
\item
  \textbf{Prompts and sample.} Ten flash-fiction prompts in the main
  study, twelve in the follow-up, 200 judged pairs per judge. The prompt
  is the unit of inference, so intervals rest on ten clusters, and a
  percentile bootstrap over so few clusters can understate uncertainty;
  the drift bound may be optimistic for that reason.
\item
  \textbf{Unfinished stories in the main study.} Stories were cut at 220
  new tokens and the stopping reason was not recorded. Some main-study
  stories end mid-sentence, and judges saw them as written. The cut
  falls on the longest stories, so it is tangled with the judges'
  preference for length (Section 7); we did not separate the two.
\item
  \textbf{The ``better'' probe's wording.} Every ``better'' result used
  ``The more better response is Response''. Every stage passed the
  competence check with it. The grammatical version, run on the two SFT
  judges, also passes (.95 and .97 on shuffled pairs); the other five
  stages and the main measurement were not re-run with it.
\item
  \textbf{Prompt fit.} No judge reliably detects an off-prompt story
  (.50 to .70 under ``better''), so our claims concern the text, not its
  fit to the prompt.
\item
  \textbf{Self-scoring at the ends of the family's view.} Base-stage and
  final-stage stories are scored partly by their own writer, which
  widens the gap between the end stages by construction (Section 5). The
  judge analysis inherits a milder form, since its surprise score for a
  base-stage story comes from the model that wrote it. Scoring surprise
  with the judge's own family instead, which involves no self-scoring,
  leaves the ``creative'' conclusion unchanged (no trained judge prefers
  the predictable story more than its base does), though Zephyr's DPO
  slope shrinks from +.120 to +.029. Under ``better'' it moves the drift
  bounds both ways: OLMo-2's shrink slightly, and Zephyr's DPO bound
  grows from 0.5 to 2.1 points (Section 6).
\item
  \textbf{Sign of the judge slope.} Pairs cross training stages, so the
  direction of any judge preference is confounded with stage (Section
  6). We claim only that the predicted drift is not detected, with a
  post hoc bound on its size.
\item
  \textbf{No judge read its own family.} Every judge read the other
  family's stories, so this study tests whether training gives a judge a
  taste for predictable writing. It does not test whether a judge favors
  its own family's stories, the self-preference that Panickssery et
  al.~(2024) and Li et al.~(2025) study.
\item
  \textbf{No human judgment of creativity in the main study.} ``Better''
  is the judge's own quality axis. It could be polish rather than value.
  Its correlation with the predictability gap stayed between .11 and .29
  in absolute value at every stage, under the .5 threshold we set in
  advance for treating it as entangled.
\item
  \textbf{The follow-up.} One writer, two raters (the authors) who
  scored independently and resolved disagreements by discussion into one
  agreed score per story, so no inter-rater agreement is on record, eight pairs (four natural pairs, the widest by design, and
  four sampled heated pairs), and half-point scores. No statistical test
  is claimed from the blind reads, and the localization check covers three stories.
\end{itemize}

\subsection{10. Conclusion}\label{conclusion}

We asked whether a model's judgment drifts with its writing as training
proceeds. For OLMo-2 and Zephyr, the writing drifts. Every trained stage
writes more predictably than its base, and each family comes to find its
own writing familiar. The judgment barely moves toward predictable
writing: asked which story is better, no trained judge's preference for
predictable stories grows by as much as a point of pick probability at
its estimate, and a one-sided 95\% upper
bound on that growth is about 5 points on the widest pairs. Training instead makes judges more
swayed by length and position, and it breaks ``creative'' as a question,
so that trained judges asked for creativity no longer reliably prefer a
story to its own scrambled words. The cleaner test that could settle the
sign of a judge's taste needs less predictable stories that reliably
hold together. Pushing the Zephyr writer toward them broke too many to
pass a quality floor, which is a warning for anyone who reads low
predictability as novelty.

\subsection{Appendix A. Where the reader's surprise
sits}\label{appendix-a.-where-the-readers-surprise-sits}

If a high surprise score meant a broken story, the broken sentences
themselves should be the most surprising ones. We checked that in the
three broken stories. Before any sentence was scored, the break
sentences in the three logic-broken stories were marked from the texts
alone, seven sentences in all, and the marks were committed. Then each
sentence was scored for average surprise. One of the seven break
sentences fell in its story's most surprising quarter, where chance
would put about two, and four of the seven sat above their story's
median sentence. By the reading rule written with the marks, that
pattern is diffuse, so the breaks are not where the reader's surprise
concentrates. Among the most surprising sentences were openings and odd
premises: ``a bill for over a thousand cosmic surgeries'', ``a deep blue
river with no name'', a clock that ``blinked 12:00 in neon red''. Three
stories cannot carry a claim, and an opening sentence has the least
context, so part of its surprise is position.

\subsection{Author contributions}\label{author-contributions}

A.N.K.: conceptualization, methodology, software, investigation, formal analysis, data curation,
writing – original draft, writing – review \& editing. A.B.: investigation (blind ratings),
writing – review \& editing. Both authors read and approved the final manuscript.

\subsection{AI use}\label{ai-use}

AI tools were used to assist with code development and to improve the clarity and readability of the
text. The study was conceived, designed and directed by the authors. The first author formulated the
research question, designed the experiments and their pre-registered decision rules, performed the
blind quality ratings and the analyses, interpreted the results, drafted the manuscript and revised
it critically. The second author contributed to the blind ratings and critically revised the
manuscript. Both authors read and approved the final version and take full responsibility for its
content.

\subsection{Code and data}\label{code-and-data}

Code, generated stories, predictability scores, judge outputs and the
dated design documents are at
https://github.com/IamArmanNikkhah/judge-is-not-its-twin.

\subsection{References}\label{references}

Chakrabarty, T., Laban, P., Agarwal, D., Muresan, S., and Wu, C.-S.
(2024). Art or artifice? Large language models and the false promise of
creativity. CHI 2024. arXiv:2309.14556.

Dubois, Y., Galambosi, B., Liang, P., and Hashimoto, T. B. (2024).
Length-controlled AlpacaEval: A simple way to debias automatic
evaluators. arXiv:2404.04475.

Gu, J., Jiang, X., Shi, Z., Tan, H., Zhai, X., et al.~(2024). A survey on
LLM-as-a-judge. arXiv:2411.15594.

Hada, R., Gumma, V., de Wynter, A., Diddee, H., Ahmed, M., Choudhury,
M., Bali, K., and Sitaram, S. (2024). Are large language model-based
evaluators the solution to scaling up multilingual evaluation? EACL
Findings 2024. arXiv:2309.07462.

Holtzman, A., Buys, J., Du, L., Forbes, M., and Choi, Y. (2020). The
curious case of neural text degeneration. ICLR 2020. arXiv:1904.09751.

Jiang, A. Q., Sablayrolles, A., Mensch, A., et al.~(2023). Mistral 7B.
arXiv:2310.06825.

Karouzos, C., Tan, X., and Aletras, N. (2026). Where does output
diversity collapse in post-training? arXiv:2604.16027.

Li, D., Sun, R., Huang, Y., Zhong, M., Jiang, B., et al.~(2025).
Preference leakage: A contamination problem in LLM-as-a-judge.
arXiv:2502.01534.

Mohammadi, B. (2024). Creativity has left the chat: The price of
debiasing language models. arXiv:2406.05587.

Panickssery, A., Bowman, S. R., and Feng, S. (2024). LLM evaluators
recognize and favor their own generations. arXiv:2404.13076.

Peeperkorn, M., Kouwenhoven, T., Brown, D., and Jordanous, A. (2024). Is
temperature the creativity parameter of large language models?
arXiv:2405.00492.

Rafailov, R., Sharma, A., Mitchell, E., Ermon, S., Manning, C. D., and
Finn, C. (2023). Direct preference optimization: Your language model is
secretly a reward model. NeurIPS 2023. arXiv:2305.18290.

Runco, M. A., and Jaeger, G. J. (2012). The standard definition of
creativity. Creativity Research Journal, 24(1), 92-96.
doi:10.1080/10400419.2012.650092.

Team OLMo, Walsh, P., Soldaini, L., Groeneveld, D., Lo, K., et
al.~(2024). 2 OLMo 2 Furious. arXiv:2501.00656.

Tunstall, L., Beeching, E., Lambert, N., Rajani, N., Rasul, K., et
al.~(2023). Zephyr: Direct distillation of LM alignment.
arXiv:2310.16944.

Tutone, A., Franceschelli, G., and Musolesi, M. (2026). The limits of
automatic evaluation of creativity in large language models.
arXiv:2608.23705.

Wang, P., Li, L., Chen, L., Cai, Z., Zhu, D., et al.~(2023). Large
language models are not fair evaluators. arXiv:2305.17926.

Wataoka, K., Takahashi, T., and Ri, R. (2024). Self-preference bias in
LLM-as-a-judge. arXiv:2410.21819.

Zhang, H., Duckworth, D., Ippolito, D., and Neelakantan, A. (2020).
Trading off diversity and quality in natural language generation.
arXiv:2004.10450.

Zhang, J., Yu, S., Chong, D., Sicilia, A., Tomz, M. R., et al.~(2025).
Verbalized sampling: How to mitigate mode collapse and unlock LLM
diversity. arXiv:2510.01171.

Zheng, L., Chiang, W.-L., Sheng, Y., Zhuang, S., Wu, Z., et al.~(2023).
Judging LLM-as-a-judge with MT-Bench and Chatbot Arena. NeurIPS 2023
Datasets and Benchmarks. arXiv:2306.05685.

Zhou, H., Huang, H., Zhang, R., Chen, K., Xu, B., et al.~(2026). Toward
robust LLM-based judges: Taxonomic bias evaluation and debiasing
optimization. arXiv:2603.08091.

\end{document}